%% file: root.tex
\documentclass[letterpaper, 10pt, conference]{ieeeconf}      

\usepackage{times}
\usepackage{epsfig}
\usepackage{graphicx}
\usepackage{amsmath}
\usepackage{amssymb}
\usepackage{textcomp}
\usepackage{multirow}
\usepackage{adjustbox}

\usepackage{multirow}
\usepackage{colortbl}
\usepackage{hhline}
\usepackage{subcaption}
\usepackage{verbatim}

\usepackage{gensymb}
\usepackage{xcolor}

\makeatletter
\let\NAT@parse\undefined
\DeclareRobustCommand\onedot{\futurelet\@let@token\@onedot}
\def\@onedot{\ifx\@let@token.\else.\null\fi\xspace}

\makeatother

\usepackage[colorlinks,pagebackref=false,citecolor=blue,bookmarks=false,hypertexnames=true]{hyperref}

\IEEEoverridecommandlockouts                              

\title{\LARGE \bf
BEV-MODNet: Monocular Camera based Bird's Eye View\\  Moving Object Detection  for Autonomous Driving 
}

\author{Hazem Rashed$^{1}$, Mariam Essam$^{1}$, Maha Mohamed$^{1}$, Ahmad El Sallab$^{1}$ and Senthil Yogamani$^{2}$ \\ %
$^{1}$Valeo R\&D Cairo, Egypt \quad\quad $^{2}$Valeo Visions Systems, Ireland \\
         {\tt \small\{hazem.rashed, mariam.essam, maha.mohamed, ahmad.el-sallab, senthil.yogamani\}@valeo.com}
}

\begin{document}

\maketitle
\thispagestyle{empty}
\pagestyle{empty}

\input{include/abstract.tex}

\input{include/introduction.tex}

\input{include/RelatedWork.tex}

\input{include/methodology.tex}

\input{include/experiments.tex}

\input{include/results.tex}

\input{include/conclusions.tex}






\bibliographystyle{ieee}
\bibliography{references/egbib}

\input{include/supplement.tex}
\end{document}

%% file: include/abstract.tex
\begin{abstract}
Detection of moving objects is a very important task in autonomous driving systems. After the perception phase, motion planning is typically performed in Bird's Eye View (BEV) space. This would require projection of objects detected on the image plane to top view BEV plane. Such a projection is prone to errors due to lack of depth information and noisy mapping in far away areas. CNNs can leverage the global context in the scene to project better. In this work, we explore end-to-end Moving Object Detection (MOD) on the BEV map directly using monocular images as input. To the best of our knowledge, such a dataset does not exist and we create an extended KITTI-raw dataset consisting of 12.9k images with annotations of moving object masks in BEV space for five classes. The dataset is intended to be used for class agnostic motion cue based object detection and classes are provided as meta-data for better tuning. 
We design and implement a two-stream RGB and optical flow fusion architecture which outputs motion segmentation directly in BEV space. We compare it with inverse perspective mapping of state-of-the-art motion segmentation predictions on the image plane.  We observe a significant improvement of 13\% in mIoU using the simple baseline implementation. This demonstrates the ability to directly learn motion segmentation output in BEV space. Qualitative results of our baseline and the dataset annotations can be found in \url{https://sites.google.com/view/bev-modnet}. 

\end{abstract}

%% file: include/introduction.tex
\section{INTRODUCTION}

Moving object detection has gained significant attention recently especially for autonomous driving applications \cite{horgan2015vision}. Motion information can be used as a signal for class-agnostic detection. For example, current systems come with appearance based vehicle and pedestrian detectors. They won't be able detect unseen classes like animals which can cause accidents. Motion cues can be used to detect any moving object regardless of its class, and hence the system can use it to highlight unidentified risks. Moving objects also need to be detected for their removal in SLAM systems \cite{tripathi2020trained}.

Sensor fusion is typically used to obtain an accurate and robust perception. A common representation for all sensors fusion is the BEV map which defines the location of the objects relative to the ego-vehicle from top-view perspective. 
BEV maps also provides a better representation than image view  as they minimize the occlusions between objects that lie on the same line of sight with the sensor. In case of visual perception on image view, a projection function is applied to map them to the top-view BEV space.

Such a projection is usually error prone  due to the absence of depth information. Deep learning on the other hand can be used to improve this inaccuracy by learning the objects representation directly in BEV representation. There has been efforts to explore deep learning performance for BEV object detection using camera sensor and there has been also efforts in motion segmentation on front view. However, there is no literature in end-to-end learning of BEV motion segmentation. In this work, we attempt to tackle such limitation through the following contributions: 

\begin{itemize}
    \item We create a dataset comprising of 12.9k images containing BEV pixel-wise annotation for moving and static vehicles for 5 classes.
    \item We design and implement a simple end-to-end baseline architecture demonstrating reasonable performance. 
    \item We compare our results against conventional Inverse Perspective Mapping (IPM) \cite{mallot1991inverse} approach and show a significant improvement of over 13\%.
\end{itemize}

\begin{figure}[!t]
\centering
\includegraphics[width=\columnwidth]{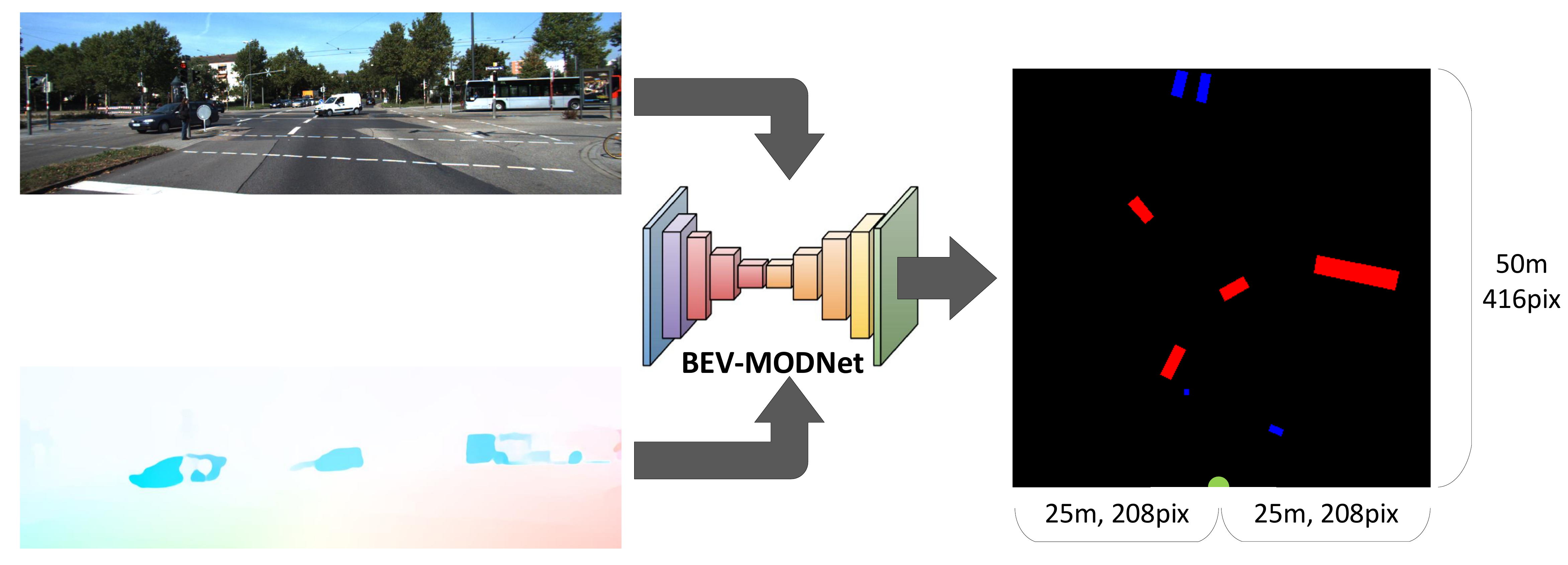}
\caption{Our model predicts bird's eye view motion segmentation using RGB image and optical flow. Red and blue regions denote moving and static vehicles. The green circle shows the ego-vehicle position.} 
\vspace{-0.51cm}
\label{fig:CartesianPC}
\end{figure}

The paper is organized as follows. Section \ref{sec:relatedwork} reviews the related work in MOD task. Section \ref{sec:methodology} discusses our proposed dataset and baseline architecture and its implementation. Section \ref{sec:experiments} describes the experimental setup and analysis of our results. Finally, section \ref{sec:conc} provides the final conclusion.

%% file: include/RelatedWork.tex
\section{RELATED WORK} \label{sec:relatedwork}

\begin{figure*}[t!]
\captionsetup[subfigure]{labelformat=empty}
\centering
\begin{adjustbox}{minipage=\linewidth,scale=0.8}
\begin{subfigure}{.5\textwidth}
    \includegraphics[width=\columnwidth]{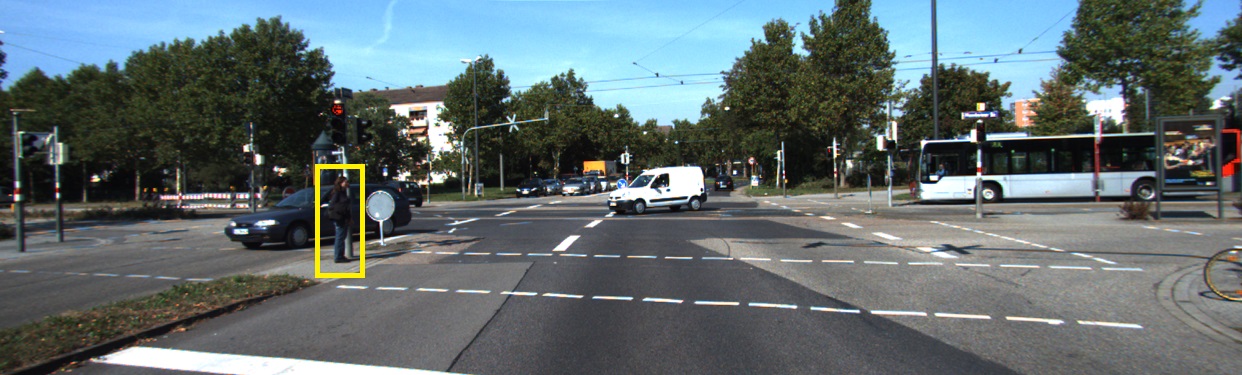}
    \vspace{-1cm}
    \caption{\textcolor{white}{(a)}}
\end{subfigure}%
\hspace{0.1cm}
\hfill
\begin{subfigure}{.5\textwidth}
    \includegraphics[width=\columnwidth]{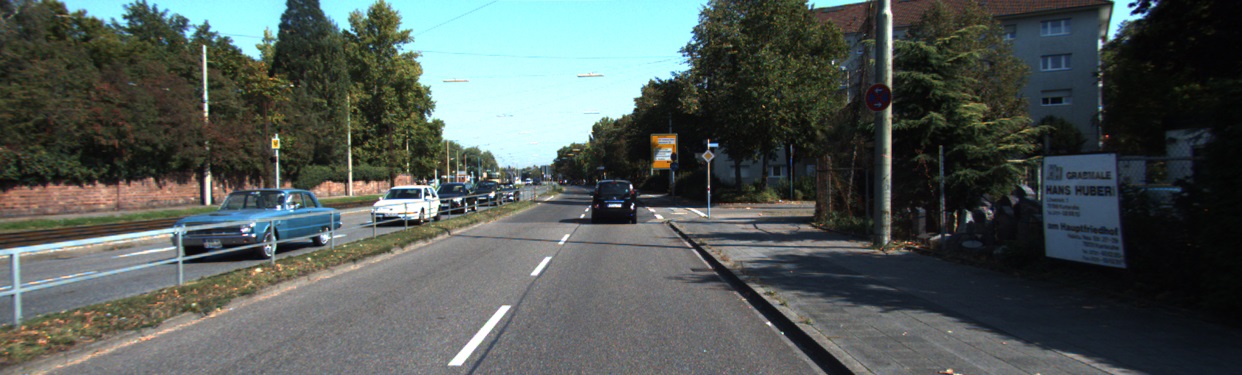}
    \vspace{-1cm}
    \caption{\textcolor{white}{(b)}}
\end{subfigure}%

    \vspace{0.1cm}

\begin{subfigure}{.5\textwidth}
    \includegraphics[width=\columnwidth]{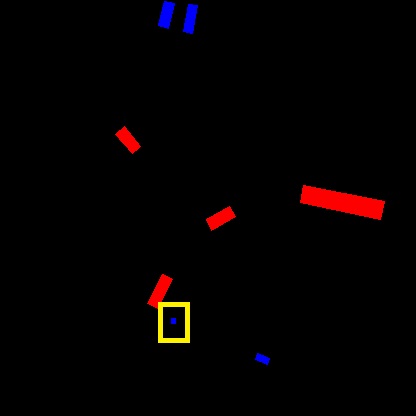}\\
    \vspace{-1cm}
    \caption{\textcolor{white}{(c)}}
\end{subfigure}%
\hspace{0.1cm}
\hfill
\begin{subfigure}{.5\textwidth}
    \includegraphics[width=\columnwidth]{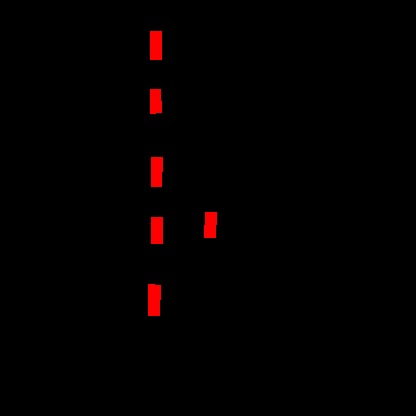}
    \vspace{-1cm}
    \caption{\textcolor{white}{(d)}}
\end{subfigure}%

\quad
\end{adjustbox}
    \caption{Few samples of our dataset. (a,b) represent KITTI RGB images. The red regions represent moving objects and the blue ones represent static ones. As demonstrated by the yellow boxes, occlusion in front view might cause difficulties for prediction, however, in BEV, occlusions are eliminated. Object instances are better separated as well reducing the need for an explicit instance segmentation.}
    \label{fig:dataset_samples}
    \vspace{-0.4cm}
  
\end{figure*}

Motion segmentation has been explored through classical approaches such as \cite{menze2015object}. Classical methods usually make use of complex algorithmic pipelines which accumulate the errors of each step to the final result providing less accuracy compared to deep learning approaches.
Foreground segmentation has been explored by \cite{jain2017fusionseg} using optical flow, however the algorithm is generic and it does not predict enough information to classify if the object is moving or static.
Video object segmentation has been explored in \cite{drayer2016object,tokmakov2017learning} using complex approaches that are not applicable to our application as they use heavy models such as R-CNN \cite{drayer2016object} and DeepLab \cite{tokmakov2017learning} which provide 8 fps only. On the other hand, \cite{siam2018modnet,siam2018real} explored moving object detection using CNNs. Appearance features are obtained from RGB images and motion features are obtained from the corresponding optical flow images which encode the scene motion. InstanceMotSeg \cite{mohamed2020monocular} extended it to obtain motion segmentation at instance level. MOD has been also explored on fisheye images in \cite{yahiaoui2019fisheyemodnet} using wide angle cameras and higher distortion levels relative to conventional images. The approach has been evaluated on \cite{yogamani2019woodscape} dataset which provides fisheye images from 4 surround-view cameras and their corresponding MOD annotations captured from real AD scenes.

In a typical AD pipeline, planning and prediction are done on a top-view map, where height information is usually discarded due to its low importance relative to to BEV information. Many recent algorithms attempted to explore environment perception on BEV such as \cite{casas2018intentnet, bansal2018chauffeurnet, houston2020one}. Most of this work focus mainly on scenes obtained by LiDAR sensor which is an expensive sensor for commercial vehicles to deploy.  Time-of-flight (ToF) sensors such as LiDAR provide depth information which makes projection of the scene onto a BEV map relatively easier without explicit assumptions. On the other hand, due to the fact that camera is a low cost sensor which unlike LiDAR provides dense scene perception, the prediction of BEV images has recently gained a huge attention. Inverse Perspective Mapping (IPM) is a standard method to project images on BEV map. To perform such a method, 4 corresponding point pairs in a source and target frame have to be determined to compute a homography matrix for such transformation. The approach assumes flat ground surface and fixed camera extrinsic parameters which is not realistic in a lot of scenarios. Moreover, it provides noisy estimates  and breaks down in occluding scenarios. In \cite{roddick2018orthographic}, end-to-end 3D object detection from monocular images have been explored through explicit projection inside the network. The approach is computationally heavy and not suitable for real-time applications. Later, the authors expanded to semantic segmentation \cite{roddick2020predicting}. In \cite{mani2020monolayout}, an encoder-decoder architecture has been used to predict BEV directly from monocular scenes, where authors showed that CNNs are able to predict such representation without explicit projection inside the network. All the mentioned methods study BEV object detection and semantic segmentation. However, BEV motion segmentation is not explored. 


\begin{figure*}[t!]
\captionsetup[subfigure]{labelformat=empty}
\centering
\begin{adjustbox}{minipage=\linewidth,scale=0.9}
\begin{subfigure}{.245\textwidth}
    \includegraphics[width=\columnwidth]{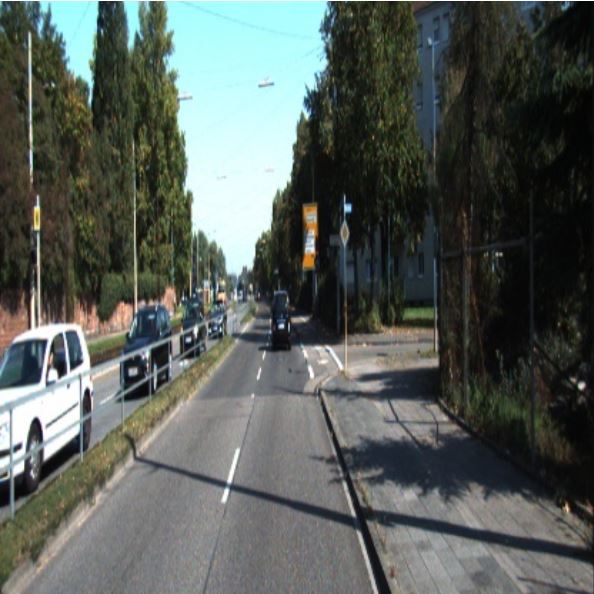}
    \vspace{-1cm}
    \caption{\textcolor{white}{(a)}}
\end{subfigure}%
\hfill
\begin{subfigure}{.245\textwidth}
    \includegraphics[width=\columnwidth]{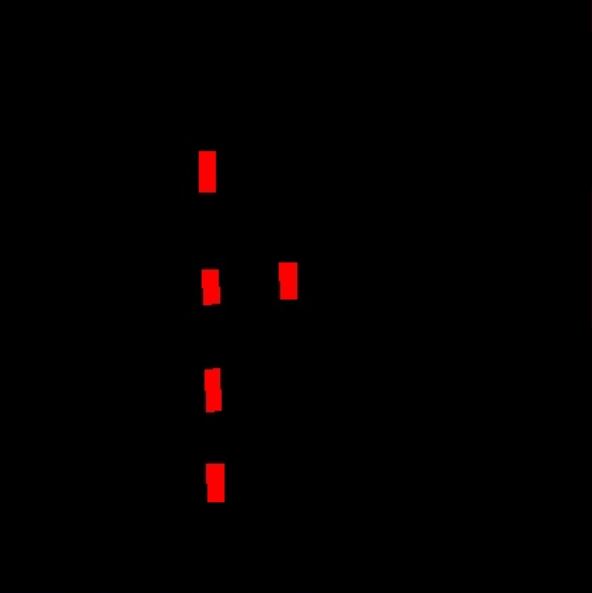}\\
    \vspace{-1cm}
    \caption{\textcolor{white}{(b)}}
\end{subfigure}%
\hfill
\begin{subfigure}{.245\textwidth}
    \includegraphics[width=\columnwidth]{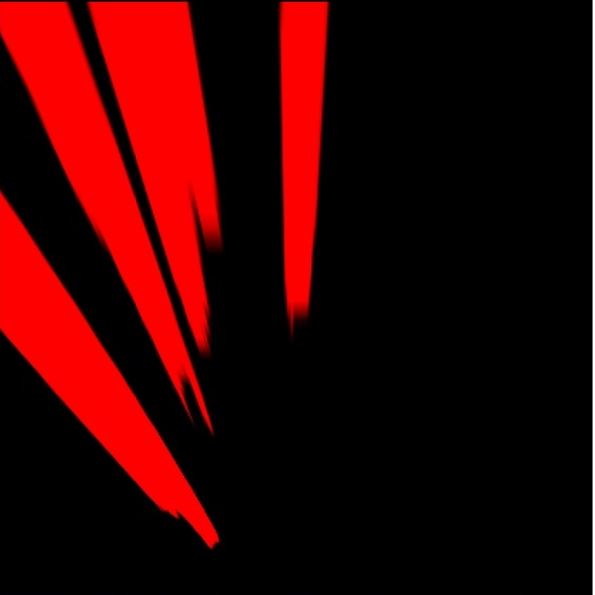}
    \vspace{-1cm}
    \caption{\textcolor{white}{(c)}}
\end{subfigure}%
\hfill
\begin{subfigure}{.245\textwidth}
    \includegraphics[width=\columnwidth]{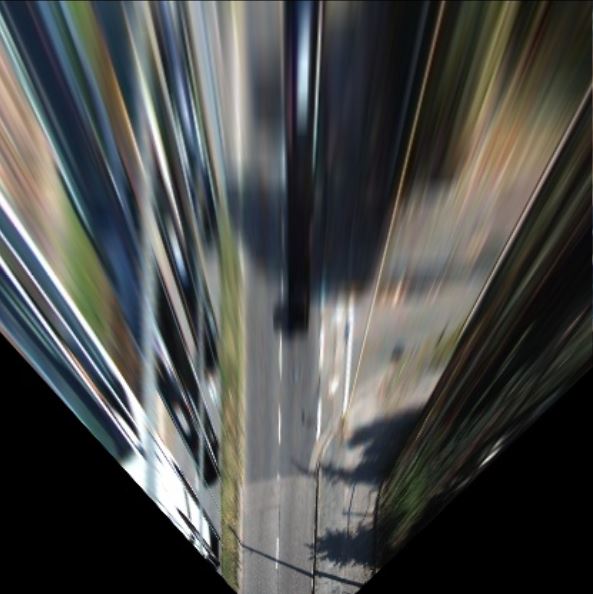}
    \vspace{-1cm}
    \caption{\textcolor{white}{(d)}}
\end{subfigure}%
\quad
\end{adjustbox}
    \caption{ Perceived objects have to be localized in a top view grid for motion planning. This can be done by either detecting in the image view and projecting to top view (c) or directly detecting in top view (b). (a) is the RGB image, (b) is motion segmentation predictions on top view, (c) is the projection of the motion segmentation on image to top view show and (d) is the projection of image to top view illustrated to better interpret (c). 
    }
    \label{fig:projection_samples}
    \vspace{-0.4cm}
  
\end{figure*}

%% file: include/methodology.tex
\section{PROPOSED METHOD} \label{sec:methodology}
In this section, we describe the proposed method including dataset generation and our network architecture.

\subsection {Inverse Perspective Mapping}
To be able to project a scene from image view into BEV, one would need 8 points to perform such operation. Four points have to be determined in the source frame and the corresponding 4 points have to be identified in the target frame. A homography matrix is computed to transform the source image into the target image, which is usually done in an iterative approach. For the application of autonomous driving, one interesting feature would be the lanes of the road. They can be used so that they become parallel to each other in the final projection. Due to the lack of 3D information, the projected image is not realistic and there is a lot of noise in the output. Figure \ref{fig:projection_samples} demonstrate a sample image of KITTI dataset when projected on BEV using IPM \cite{mallot1991inverse} in (d). We also evaluate predicting MOD using image view and then doing projection on top view using IPM in (c). As observed in the image, (c) provides very noisy output relative to (b) which is the representation we would like to learn end-to-end. To learn such representation directly using a deep network, one would need a dataset with such representation which is not available in the public datasets. Hence, we created our own dataset having MOD annotations on BEV.

\subsection{Dataset Generation}

In general, there is a limitation of large scale MOD datasets in autonomous driving. In \cite{vertens2017smsnet}, 255 images on KITTI dataset have been manually labeled for motion segmentation task. Additionally, around 3k images on Cityscapes dataset have been annotated. These numbers are relatively low, and they are only performed on image view of the camera sensor. In \cite{siam2018modnet}, 1.3k images have been weakly annotated for MOD, and it has been extended by \cite{rashed2019fusemodnet} where more KITTI sequences have been annotated for moving vehicles only. All these methods provide only image view annotation which cannot be used directly on top-view predictions. Hence, we create our own dataset which consists of 12.9k images including pixel-wise annotations for static and moving objects. Our dataset is labeled for 5 classes, and it will be released for all classes, however in our experiments we focus on training the network with \textit{vehicle} class only to simplify the problem. We choose KITTI dataset because of the extensive prior work on MOD to enable comparison. Another alternative is NuScenes \cite{caesar2020nuscenes} dataset. It is a more recent larger dataset which provides motion attribute for vehicles and pedestrians but not for cyclists and motorcyclists. 

We adapt the approach by \cite{rashed2019fusemodnet} to create our dataset, where we make use of KITTI raw sequences as they provide IMU/GPS measurements for ego-motion, LiDAR point clouds for depth and 3D boxes of the objects relative to the ego-vehicle in each frame. First, we use the IMU/GPS to compute the ego-vehicle motion in LiDAR coordinates system. We use the tracking information to compute the difference in objects positions between each two sequential frames. We project the objects into world co-ordinate system and we compute the difference between ego-vehicle and other objects motion. Using thresholding techniques, we are able to classify the surrounding objects into moving or static ones. We project the 3D points in the 3D camera co-ordinate system onto the $xz$ plane to obtain the BEV representation of the surrounding scene. 

At first, we obtained the max distance of the furthest object in the dataset and set it as maximum distance to make sure all objects are included in the dataset. However, we observe that most of the objects are closer to the ego-vehicle and most of the output maps are mostly empty because of the sparsity of very far objects. Hence, we keep the maximum distance to $50m$ to maintain reasonable resolution of the output maps. We observe that there are still false positives and negatives in our moving/static labels due to thresholding errors. We fix such errors by manual refinement of the false labels. The main task we target is motion segmentation, so we provide the annotations as pixel-wise masks for moving objects. We also provide masks for static objects which may be used to improve the motion segmentation prediction. Figure \ref{fig:dataset_samples} represents samples of our generated dataset. The blue boxes refer to static objects from BEV viewpoint and the red ones correspond to moving ones. Table \ref{tab:dataset_analysis} demonstrates details about our generated dataset. We provide annotations for 5 classes and we provide details about both static and moving objects in each class.

\begin{figure*}[!t]
\centering
\includegraphics[width=\textwidth]{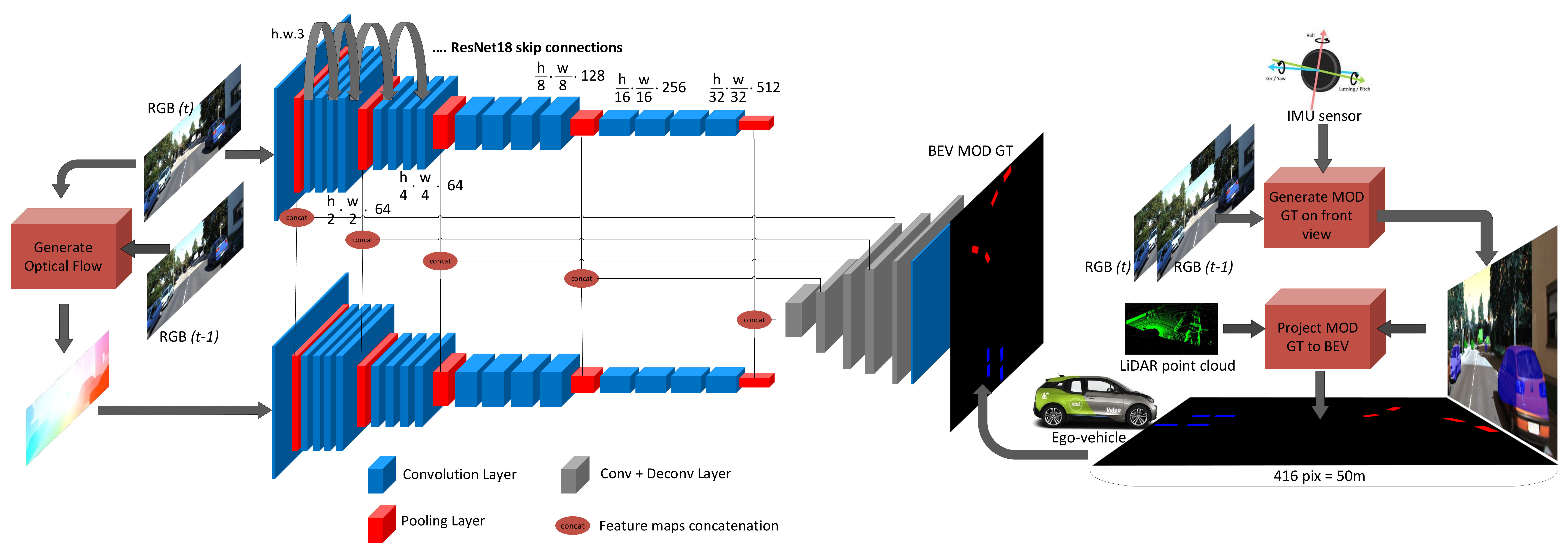}
\caption{Illustration of our BEV-MODNet architecture. Our network is fed with a single RGB image and the corresponding optical flow generated by \cite{ilg2017flownet}. We generate our MOD annotation by utilizing sequential information from KITTI dataset in addition to ego-vehicle motion information. We project our annotations on BEV view and learn BEV MOD directly.}
\vspace{-0.51cm}
\label{fig:network_arch}
\end{figure*}

\subsection{Network Architecture}
As proposed by \cite{mani2020monolayout}, an encoder-decoder architecture is able to learn a BEV representation of the surrounding scene. We follow a similar approach and make use of the architecture in \cite{kumar2021omnidet} for its high inference rate and low weight. The model in \cite{mani2020monolayout} is able to predict BEV map without focusing on moving vs static object classification using a single monocular image. On the other hand, the model in \cite{kumar2021omnidet} takes two sequential images as input and tries to understand motion implicitly. This approach has been proven by \cite{ramzy2019rst} that it provides less accuracy than feeding optical flow explicitly. We create another encoder which accepts optical flow as input and we perform feature fusion \cite{rashed2019motion} between multi-scale feature maps of both encoders and then feed the resulted tensors into the decoder. The decoder consists of 5 deconvolution layers which are preceded by convolution layers \cite{segdasvisapp19}. The final output is a binary mask which predicts a class for each pixel among the two classes (Moving and Static). 


%% file: include/experiments.tex
\section{EXPERIMENTS} \label{sec:experiments}
In this section, we provide details of our experimental setup and the analysis of the obtained results.

\subsection{Experimental Setup}
We use ResNet18 as backbone for feature extraction, where we create another encoder for capturing motion features from optical flow. We initialize our network with the ResNet18 pre-trained weights and we set the batch size to 16. The network is trained using the Ranger (RAdam\cite{liu2019variance} + LookAhead \cite{zhang2019lookahead}) optimizer. We train all the models using weighted binary cross-entropy loss function for 60 epochs. We use transposed convolution layers for upsampling purpose to finally reach the original input size. Finally, a weighted binary cross entropy loss function is used to obtain the final predictions for each pixel as a classification task among two classes, i.e, Moving and Static.

%% file: include/results.tex


\begin{figure*}[t!]
\captionsetup[subfigure]{labelformat=empty}
\centering
\begin{adjustbox}{minipage=\linewidth,scale=1}
\begin{subfigure}{.245\textwidth}
    \includegraphics[width=\columnwidth]{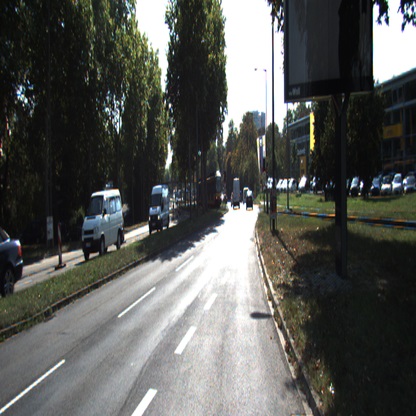}
    \vspace{-1cm}
    \caption{\textcolor{white}{(a)}}
\end{subfigure}%
\hspace{0.01cm}
\begin{subfigure}{.245\textwidth}
    \includegraphics[width=\columnwidth]{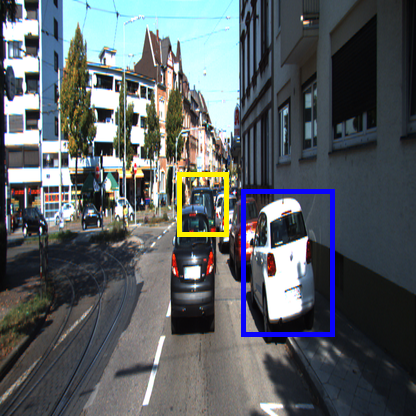}\\
    \vspace{-1cm}
    \caption{\textcolor{white}{(b)}}
\end{subfigure}%
\hspace{0.01cm}
\begin{subfigure}{.245\textwidth}
    \includegraphics[width=\columnwidth]{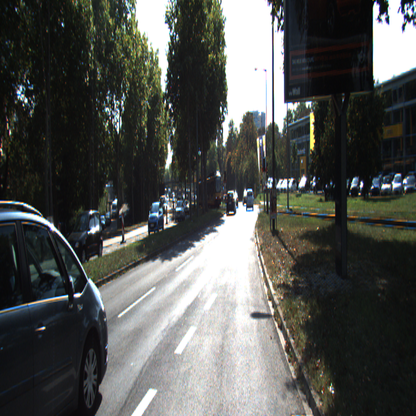}
    \vspace{-1cm}
    \caption{\textcolor{white}{(c)}}
\end{subfigure}%
\hspace{0.01cm}
\begin{subfigure}{.245\textwidth}
    \includegraphics[width=\columnwidth]{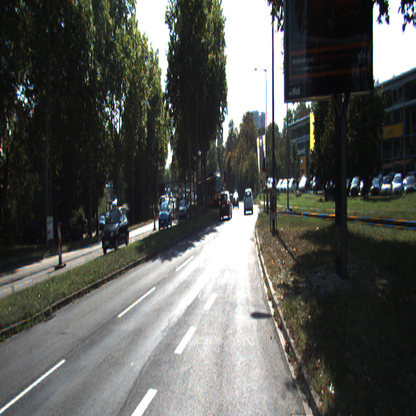}
    \vspace{-1cm}
    \caption{\textcolor{white}{(d)}}
\end{subfigure}%

\vspace{0.08cm}

\begin{subfigure}{.245\textwidth}
    \includegraphics[width=\columnwidth]{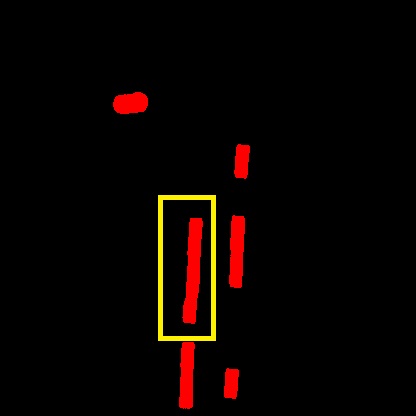}
    \vspace{-1cm}
\end{subfigure}%
\hspace{0.01cm}
\begin{subfigure}{.245\textwidth}
    \includegraphics[width=\columnwidth]{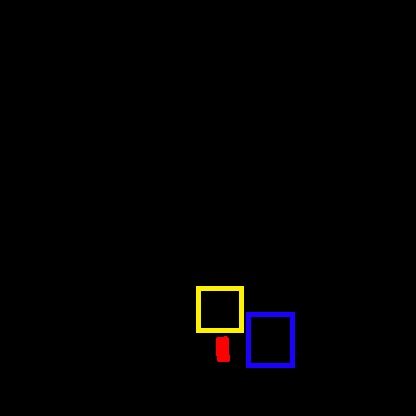}\\
    \vspace{-1cm}
\end{subfigure}%
\hspace{0.01cm}
\begin{subfigure}{.245\textwidth}
    \includegraphics[width=\columnwidth]{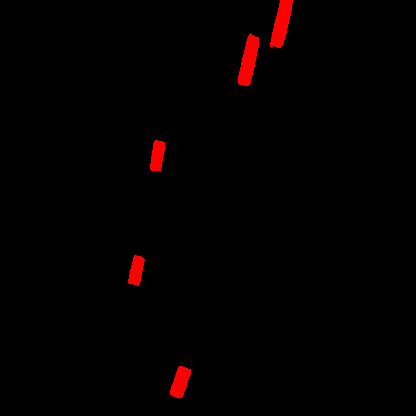}
    \vspace{-1cm}
\end{subfigure}%
 \hspace{0.01cm}
\begin{subfigure}{.245\textwidth}
    \includegraphics[width=\columnwidth]{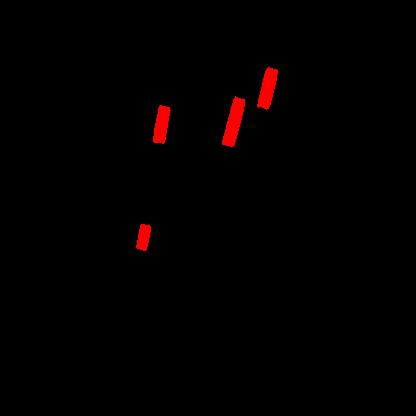}
    \vspace{-1cm}
\end{subfigure}%

\vspace{0.64cm}

\begin{subfigure}{.245\textwidth}
    \includegraphics[width=\columnwidth]{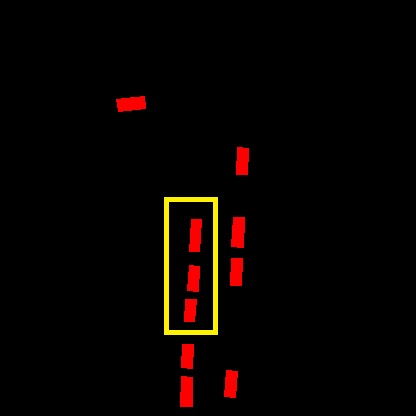}
    \vspace{-1cm}
\end{subfigure}%
\hspace{0.01cm}
\begin{subfigure}{.245\textwidth}
    \includegraphics[width=\columnwidth]{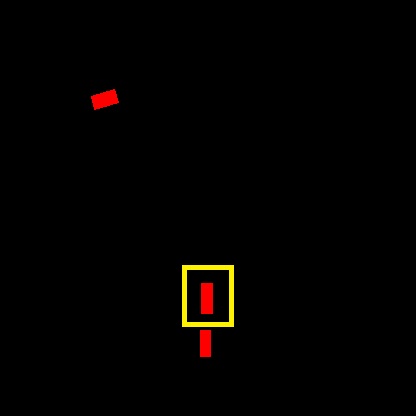}\\
    \vspace{-1cm}
\end{subfigure}%
\hspace{0.01cm}
\begin{subfigure}{.245\textwidth}
    \includegraphics[width=\columnwidth]{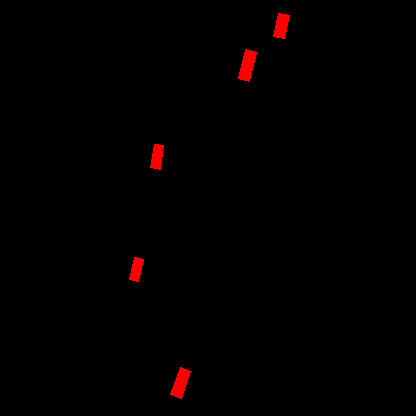}
    \vspace{-1cm}
\end{subfigure}%
\hspace{0.01cm}
\begin{subfigure}{.245\textwidth}
    \includegraphics[width=\columnwidth]{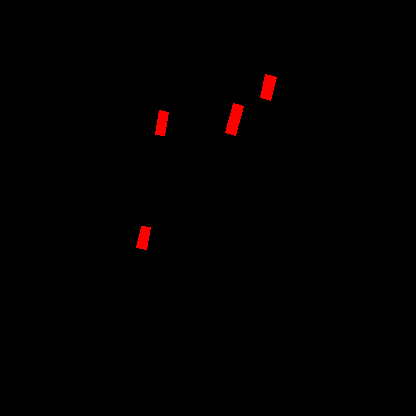}
    \vspace{-1cm}
\end{subfigure}%
 \vspace{0.5cm}
\quad
\end{adjustbox}
    \caption{Qualitative results of our baseline two-stream RGB and optical flow network which predicts motion segmentation directly on BEV. First row corresponds to RGB images, second row corresponds to predictions and third row corresponds to the ground truth. First two columns show challenging scenarios and the last two columns show easy scenarios where the model performed well.
    Yellow boxes in column (a) illustrate how nearby objects are merged in top view segmentation. Column (b) illustrates missing segmentation of the object in yellow probably due to occluded footpoint. Static object in blue box was correctly suppressed. }
    \label{fig:qualitative}
    \vspace{-0.4cm}
  
\end{figure*}


\subsection{Results}
Table \ref{tab:quantitative} demonstrate the results using our baseline network to predict BEV MOD end-to-end vs doing the prediction on Front view and performing IPM afterwards. We evaluate both predictions vs our generated ground truth and we obtain significant improvement over IPM approach by approx 7\% in mIoU.
Figure \ref{fig:qualitative} demonstrates the output of our network in 2nd row vs the ground truth from our dataset in 3rd row. It is shown that CNN is able to predict BEV directly through an encoder-decoder architecture as proposed by \cite{mani2020monolayout}. However, the network is not able to distinguish between vehicles in some cases as described in the first column. In the second column, the object highlighted in yellow is very hard to predict where most of the object is occluded behind the front vehicle. On the other hand, the static object highlighted in blue has been suppressed correctly which shows the importance of optical flow to capture the motion information in the scene. The third and fourth columns demonstrate our results for multiple objects in the scene. Results show decent output where the model can be used a baseline for benchmarking. 
However, the length of some objects are not captured perfectly and some of the objects are mis-classified as false static objects, which shows that there is still room for improvement. Perhaps one approach to explore is to define an explicit learning-based BEV projection model inside the network to overcome such inaccuracies. 

\begin{table}[t]
\caption{ Class distribution of moving and static objects in our dataset.
}
\centering
\resizebox{0.45\textwidth}{!}{
\begin{tabular}{|l|l|c|c|c|c|c|}
\hline
Type/Class & Car & Truck & Van  & Pedestrian & Cyclist  \\ \hline
Static & 28001 & 323 & 2984 & 920 & 177 \\
Moving & 8527 & 982 & 1410 & 1356 & 1301\\
Total & 36528 & 1305 & 4394 & 2276 & 1478\\
\hline
\end{tabular}
}
\label{tab:dataset_analysis}
\end{table}

\begin{table}[t]
\centering
\caption{Quantitative comparison of different approaches.} 
\resizebox{0.5\textwidth}{!}{
\begin{tabular}{|l||l|l|}
\hline
Experiment & mIoU & fps\\ \hline
\{RGB + Optical Flow\} + IPM re-projection & 47.9 & 73 \\ \hline
\{RGB + RGB (prev)\} end-to-end BEV output & 53.3 & 85 \\ \hline
\{RGB + Optical Flow\} end-to-end BEV output & \textbf{54.5} & \textbf{85} \\ \hline
\end{tabular}
}
\label{tab:quantitative}

\vspace{3mm}

\caption{Ablation study of the effect of accuracy on detection range.} 
\resizebox{0.24\textwidth}{!}{
\begin{tabular}{|l||l|}
\hline
{\centering Detection Range} & mIoU\\ \hline
 0-10m & 51.8 \\ \hline
 10-20m & 53.5 \\ \hline
 20-30m & 55.2 \\ \hline
 30-40m & 55.8 \\ \hline
 40-50m & 53.8 \\ \hline

\end{tabular}
}
\label{tab:ablation_acc_vs_depth}
\end{table}

We also observe that accuracy of the predictions decrease with increasing the depth from the camera sensor. To evaluate that quantitatively, we divide the view range into 5 bins, 10m each, and we compute mIoU over all the images for the 5 bins separately, where the results are tabulated in Table \ref{tab:ablation_acc_vs_depth}. We observe that very close objects are not captured entirely and this is expected because very close moving objects are usually moving in the same direction with almost same speed of the ego-vehicle such as passing vehicles. Due to motion parallax problem, such vehicles appear almost fixed relative to the ego-vehicle and hence very hard to detect. Accuracy of the model reaches its maximum in the middle ranges from 10m to 30m from the camera sensor and then decrease again as we go far away from the sensor. This is intuitive because as we go far away from the camera sensor, the vehicles appear smaller and moving slowly from the viewpoint of the ego-vehicle. This makes the optical flow vectors associated with such vehicles are small compared to the closer vehicles, and hence they are harder to detect. Moreover, due to absence of depth sensor, ambiguity increases with increased depth which might cause inaccurate predictions. We also observe that using the same input modalities and same architecture, the network learns motion segmentation in front view in a better accuracy. This is expected because in BEV approach the network learns an additional task to motion segmentation which is BEV projection. However, overall, BEV end-to-end learning provides better accuracy than learning in front view then projecting using IPM as demonstrated in Table \ref{tab:quantitative}. Using more complex models than our model, better representation can be learnt using our dataset with higher accuracy.


%% file: include/conclusions.tex
\section{CONCLUSION} \label{sec:conc}
In this work, we explore the idea of learning moving object detection directly in BEV space. We create a dataset that consists of 12.9k images having annotations for MOD on 5 classes. We design a deep network to predict such representation directly and we compare our results with standard IPM approach, where we show a significant improvement of 13\% in mIoU. However, our qualitative results illustrate that there are significant gaps and more research is needed to improve the performance compared to our simple baseline. Thus, we release the dataset publicly and we hope it motivates further research in class agnostic moving object detection.